%% file: main.tex
\documentclass[10pt,twocolumn,letterpaper]{article}

\usepackage[pagenumbers]{cvpr} 


\usepackage{colortbl}

\definecolor{cvprblue}{rgb}{0.21,0.49,0.74}
\usepackage[pagebackref,breaklinks,colorlinks,allcolors=cvprblue]{hyperref}

\usepackage[capitalize]{cleveref}
\crefname{section}{Sec.}{Secs.}
\Crefname{section}{Section}{Sections}
\Crefname{table}{Table}{Tables}
\crefname{table}{Tab.}{Tabs.}

\input{definitions}

\def\paperID{3} 
\def\confName{CVPR}
\def\confYear{2025}

\title{Towards Texture- And Shape-Independent 3D Keypoint Estimation in Birds}

\author{Valentin Schmuker$^{1}$, Alex Hoi Hang Chan$^{2,3,4}$, Bastian Goldluecke$^{1,2}$, Urs Waldmann$^{1,2}$\\ \\
\normalsize $^1$Department of Computer and Information Science, University of Konstanz, Germany \\
\normalsize $^2$Centre for the Advanced Study of Collective Behaviour,  University of Konstanz, Germany\\
\normalsize $^3$Department of Collective Behavior, Max Planck Institute of Animal Behavior, Germany\\
\normalsize $^4$Department of Biology, University of Konstanz, Germany \\
\small urs.waldmann@uni-konstanz.de
}

\begin{document}
\maketitle
\input{sec/0_abstract}
\input{sec/1_intro}
\input{sec/2_related}
\input{sec/3_technical}
\input{sec/4_evaluation}
\input{sec/5_conclusion}
\input{sec/6_acknowledgements}
{
    \small
    \bibliographystyle{ieeenat_fullname}
    \bibliography{main}
}


\end{document}

%% file: definitions.tex
\newcommand{\subheading}[1]{\textbf{#1}.}

\newcolumntype{a}{>{\columncolor{verylightgray}}c}

\definecolor{verylightgray}{HTML}{E0E0E0}


%% file: sec/0_abstract.tex
\begin{abstract}
In this paper, we present a texture-independent approach to estimate and track 3D joint positions of multiple
pigeons.
For this purpose, we build upon the existing 3D-MuPPET framework, which estimates and tracks the 3D poses of up to $10$ pigeons using a multi-view camera setup.
We extend this framework by using a segmentation method that generates silhouettes of the individuals, which are then used to estimate 2D keypoints.
Following 3D-MuPPET, these 2D keypoints are triangulated to infer 3D poses, and identities are matched in the first frame and tracked in 2D across subsequent frames.
Our proposed texture-independent approach achieves comparable accuracy to the original texture-dependent 3D-MuPPET framework.
Additionally, we explore our approach's applicability to other bird species.
To do that, we infer the 2D joint positions of four bird species without additional fine-tuning the model trained on pigeons and obtain preliminary promising results.
Thus, we think that our approach serves as a solid foundation and inspires the development of more robust and accurate texture-independent pose estimation frameworks.
\end{abstract}

%% file: sec/1_intro.tex
\section{Introduction}
%
\input{figures/Qualitative_results}
Pose estimation and tracking are among the major challenges in computer vision for animals.
Recent advances in deep learning have led to the development of several frameworks that enable precise tracking and pose estimation~\cite{Mathis:2018,Xu:2022,waldmann2022improving,giewald2022nepu}, significantly aiding research in areas like psychology~\cite{Diaz-Rojas:2024}, robotics~\cite{Islam:2021} and biology~\cite{Waldmann:2024,Chimento:2024}.
Especially for the research of collective animal behavior, such frameworks are of great importance.
The accurate posture estimation and tracking of collectives can provide researchers with deeper insights into the causes, structures and dynamics of animal behavior~\cite{Koger:2023,Couzin:2023,nagy2023smart}.

However, deep neural networks present their challenges because they require massive amounts of high-quality training data.
The lack of annotated training datasets often limits the applicability and generalizability of those models, especially in 3D.
The creation of
training datasets are typically
a very labor-intensive and therefore expensive option.

A key challenge in animal pose estimation is to ensure model generalization across species and environments.
Initiatives like SuperAnimal~\cite{Ye:2024}
demonstrate that models can generalize effectively if trained on diverse and extensive datasets.
Ye et al.~\cite{Ye:2024} compile their dataset by combining multiple existing datasets, a particularly effective approach for animal classes with numerous available datasets,
like
rodents and quadrupeds. In contrast, for less-represented classes like birds, fewer datasets are available, limiting the generalizability that can be achieved from combining existing data.
%
%
%

This highlights the ongoing challenge to increase the generalization ability of posture estimation models without the need for large-scale datasets. 
This is particularly challenging because animals come in different colours and textures. 
For example, individuals can look different between different dog breeds, different species, or even within species (male vs. female birds).
Even in pigeons, individuals can have a variety of appearances, from white to darker plumage.
This introduces the challenge of having posture estimation models that can generalize to these differences.

With the recent developments of segmentation methods~\cite{Kirillov:2023} one way to address this is texture-independent pose estimation.
This approach allows the framework to be trained on data collected in laboratory environments and subsequently applied to real-world data without further tine-tuning because differences in shades and background do not affect the prediction.
Moreover, texture independence enables researchers to apply a framework across various appearances of a species, even when available training datasets only contain specific appearances.
By focusing on shape and movement rather than specific textures or colors, texture-independent methods enhance the adaptability to new appearances and species.

In this paper, we extend 3D-MuPPET~\cite{Waldmann:2024} to a texture-independent multi-animal pose estimation and tracking framework, by working on silhouette data only.
We demonstrate that the ability to reliably predict 2D and 3D poses for up to $10$ individuals remains, cf.~\cref{fig:result}.
In this way, our approach is robust against environmental changes that leave silhouettes unchanged like texture and illumination.
Furthermore, we explore the capability of our approach to generalize to new species and obtain preliminary promising results.

%% file: figures/Qualitative_results.tex

\begin{figure}
    \centering
    %
    \begin{subfigure}[b]{\columnwidth}
        \centering
        \includegraphics[width=\textwidth]{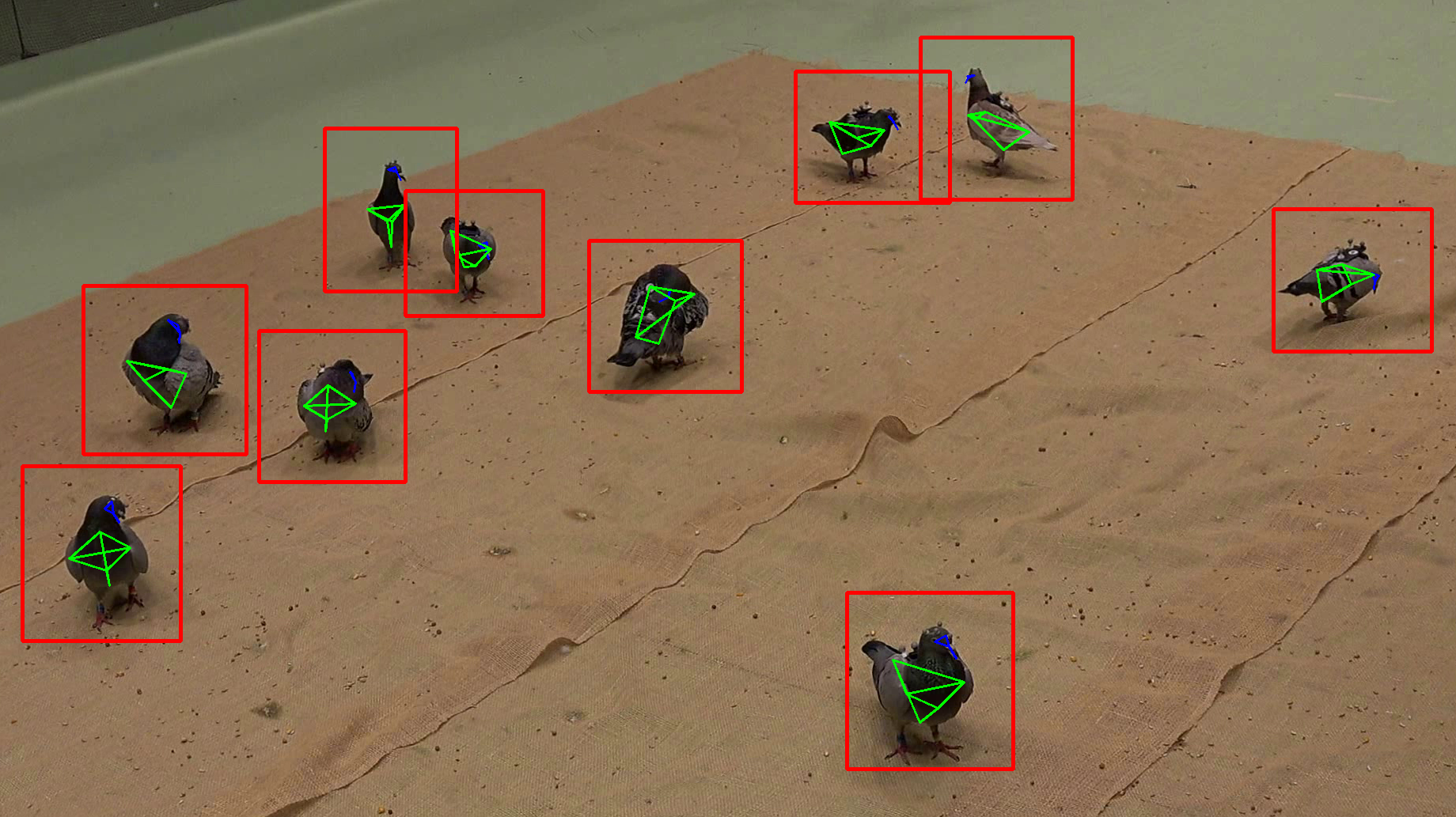}
    \end{subfigure}
    
    
    \begin{subfigure}[b]{0.495\columnwidth}
        \centering
        \includegraphics[width=\textwidth]{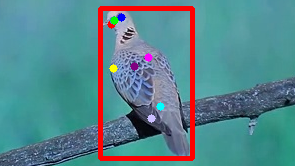}
    \end{subfigure}
    \hfill
    \begin{subfigure}[b]{0.495\columnwidth}
        \centering
        \includegraphics[width=\textwidth]{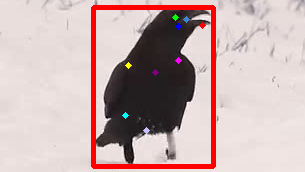}
    \end{subfigure}

    \includegraphics[width=\columnwidth]{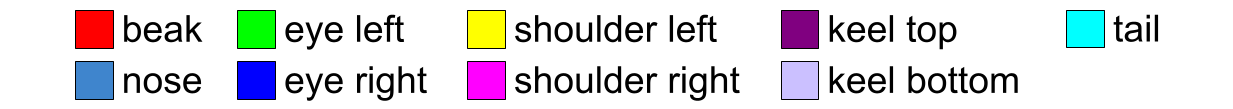}
    \caption{\textit{Qualitative results of bird poses using our texture-independent approach.}
    Top: Multi-pigeon pose estimation and tracking in 3D, projected to 2D. Green lines connect the body, blue lines the head keypoints. The detected bounding boxes are red. Example frame from 3D-POP~\cite{Naik:2023}.
    Bottom left and right: Species transfer to turtle doves and ravens, respectively.
    Detected bounding box in red. Both examples from~\cite{Ng:2022}.
    %
    %
    }
    \label{fig:result}
\end{figure}

%% file: sec/2_related.tex
\section{Related Work}
Pose estimation and tracking
is a vast field of research.
Here we review
pose estimation for animals, with a particular focus on
texture independence and
species transfer.
%

\textbf{2D Animal Pose Estimation} identifies keypoints in a single image or camera view. 
%
Methods like DeepLabCut~\cite{Mathis:2018} and LEAP~\cite{Pereira:2019} enable markerless tracking across species.
These frameworks establish pipelines for annotating 2D keypoints, which are then used to train Convolutional Neural Networks (CNNs) for keypoint prediction.
Both frameworks are extended to multiple individuals in 2D~\cite{Lauer:2022,Pereira:2022} by training on multi-animal annotations.
Our approach uses DeepLabCut~\cite{Mathis:2018} to predict 2D poses from silhouettes and then triangulates those to get 3D keypoints.
%

\subheading{3D Animal Pose Estimation}
%
\cite{Badger:2020, Biggs:2019} fit a 3D template model to the 2D silhouettes and keypoints, thereby obtaining a 3D mesh model representing the animal's pose and subsequently also the 3D keypoints.
This requires a
3D mesh model of the species under study.
Other frameworks~\cite{Waldmann:2024,Dunn:2021,Chimento:2024,joska2021acinoset,kaneko2024deciphering,han2024multi,klibaite_mapping_2025} use multi-view setups to predict 3D poses.
\cite{Dunn:2021,klibaite_mapping_2025}, e.g., employ a 3D CNN to predict 3D poses directly.
%
\cite{Waldmann:2024,joska2021acinoset} on the other hand, predict 2D poses on all camera views and triangulate these to 3D.
%
%
Our approach is similar to~\cite{Waldmann:2024}, but our 3D keypoint prediction is independent of texture and transfers to other animal species.

\subheading{Texture Independence and Species Transfer}\\
Biggs et al. \cite{Biggs:2019} introduce a texture independent 2D pose estimator for extracting 3D animal mesh models from 2D video data.
%
For
\cite{Biggs:2019} the primary motivation to develop a texture-independent system is the possibility of synthesizing training data.
Only the segmentation method
infers from real-world data, the other components can be trained using synthesized data.\\
Sun et al. \cite{Sun:2024} propose a framework,
which integrates the tasks of classification, segmentation, and animal pose estimation into a Universal Animal Perception (UniAP) model.
As a transformer-based few-shot model, UniAP~\cite{Sun:2024} is designed to perform knowledge transfer from well researched species and phenotypes to those that are less studied.
Their findings indicate that only a few annotated images are needed to apply UniAP~\cite{Sun:2024} to a new species.
In contrast to
UniAP~\cite{Sun:2024},
our approach achieves generalization through texture independence, rather than relying on additional data. Moreover, we extend
by predicting 3D keypoints.

%% file: sec/3_technical.tex
\section{Technical Framework}
First, we provide a brief description of the datasets
used in this study.
Then, we discuss 3D-MuPPET~\cite{Waldmann:2024},
followed by the changes and extensions applied to the framework.

\subsection{Datasets}
\label{sec:framework:datasets}
%
\textbf{3D-POP}~\cite{Naik:2023}
contains $59$ sequences of freely moving pigeons recorded from four views.
%
Each sequence includes either $1$, $2$, $5$, or $10$ annotated pigeons.
%
Annotations include 2D bounding boxes and nine 2D/3D keypoints (beak, nose, left/right eye, left/right shoulder, top and bottom keel and tail).
%
We use the same training, validation and test data as 3D-MuPPET~\cite{Waldmann:2024}.
%
\textbf{Wild-MuPPET}~\cite{Waldmann:2024}
includes $500$ frames of a single pigeon moving outdoors, filmed from four 4K camera views.
%
Those $500$ frames are split in an $80/20$ train test split.
\textbf{Animal Kingdom}~\cite{Ng:2022}
contains $33,099$ annotated frames
across six animal classes.
Each frame
contains one or more individuals, with only one individual being annotated with $23$ keypoints.
%
%
Because no ground truth bounding boxes are provided
we detect them with YOLOv8~\cite{Varghese:2024}.

\subsection{3D-MuPPET}
\label{sec:framework:3d-muppet}
3D-MuPPET~\cite{Waldmann:2024} is a 3D multi-pigeon pose estimation and tracking framework
that consists of a pose estimation and a tracking module.
Waldman et al.~\cite{Waldmann:2024} use three different pose estimators, while we focus on
one of them, i.e. their modified DeepLabCut~\cite{Mathis:2018},
denoted as DLC*.
%
DLC*
first uses YOLOv8~\cite{Varghese:2024} to detect and crop 
all
pigeons in each frame.
DeepLabCut~\cite{Mathis:2018} then predicts the keypoints on the
cropped images. 
To create 3D pose estimates, the 2D keypoint predictions 
of all four camera views are triangulated.
A dynamic matching algorithm based on
Huang et al.~\cite{Huang:2020} is used in the first frame to match the 2D poses of 
all views to a global ID.
In subsequent frames, each individual is tracked with SORT~\cite{Bewley:2016} in 2D, and the global ID is propagated across the trial.
%
    
\subheading{Extension}
\input{figures/Framework_extension}
%
Here, we aim to to convert the 2D pose estimator into a texture-independent one.
%
%
Hence, we use silhouettes as input.
%
Silhouettes efficiently abstract an object to its form and size, thereby minimizing the influence of lighting conditions, different phenotypes, and other factors.

We use DLC*
which consists of two parts, i.e. DLC~\cite{Mathis:2018} and YOLOv8~\cite{Varghese:2024}.
%
%
YOLOv8~\cite{Varghese:2024} is a
model that detects instances of broad categories, such as "bird", in images.
YOLOv8~\cite{Varghese:2024} is trained on a multi-object and multi-species dataset, making its classification accuracy less susceptible to changes in texture and apperance.
Therefore, YOLOv8~\cite{Varghese:2024} can be used with the same weights as in
3D-MuPPET
\cite{Waldmann:2024}.
%
Differently, we train DeepLabCut~\cite{Mathis:2018} on silhouettes to make it texture-independent.
The silhouettes are generated by creating a segmentation mask from the bounding box, during inference provided by YOLOv8~\cite{Varghese:2024}.
For segmentation, we use SAM~\cite{Kirillov:2023}
with the bounding box as a prompt.
%
%
We denote this approach
by DLCSAM.
%

We also introduce a
different approach that generates silhouettes that are more precise, cf.~\cref{fig:training}.
%
It
differs from the first in that the silhouette is further processed by isolating the largest continuous area of the mask.
%
%
We assume that the biggest continuous area of the mask corresponds to the mask of the pigeon, while the remaining areas are likely to be interference.
%
We denote this approach by DLCISO.
%

For inference on Wild-MuPPET~\cite{Waldmann:2024} we additionally introduce DLCISO-YOLO, where we use YOLOv8-seg~\cite{Varghese:2024} instead of SAM~\cite{Kirillov:2023} for segmentation. The rest of the approach remains the same as in DLCISO.
This modification leads to higher accuracy due to better masks (see~\cref{sec:evaluation:results}).
%

DLCSAM and DLCISO are trained for $100,000$ iterations.
Finally, we choose the final weights
by the smallest Root Mean Square Error (RMSE), cf.~\cref{sec:evaluation:metrics}.
%
For DLCSAM, this is after $91,000$ iterations, while for DLCISO, it is after $62,000$ iterations.

%% file: figures/Framework_extension.tex
\begin{figure}
    \begin{tabular}{c@{}c@{}c}
        \includegraphics[width=0.32\linewidth]{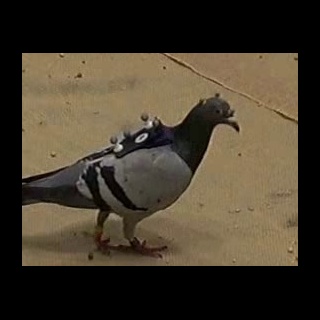}&
        \includegraphics[width=0.32\linewidth]{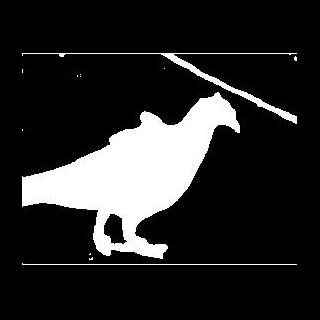}&
        \includegraphics[width=0.32\linewidth]{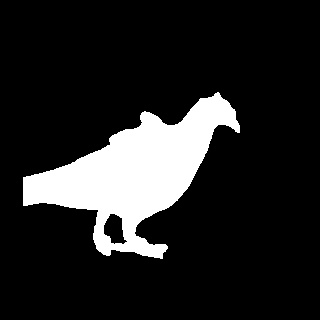}\\
        \includegraphics[width=0.32\linewidth]{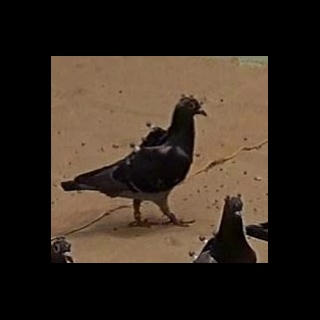}&
        \includegraphics[width=0.32\linewidth]{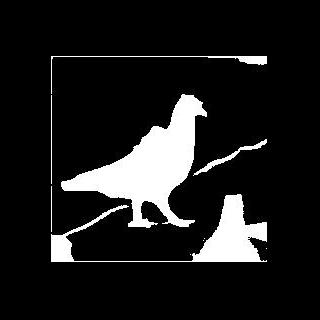}&
        \includegraphics[width=0.32\linewidth]{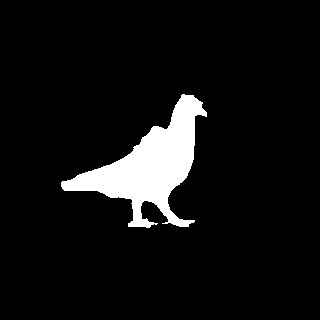}
    \end{tabular}\vspace{-10pt}
    \caption{
    %
    \textit{Texture independence.}
    %
    Two examples of our mask generation from textured images (left) using DLCSAM (middle) and DLCISO (right), cf.~\cref{sec:framework:3d-muppet}.
    %
    Example frames from~\cite{Naik:2023}.
    %
    }
    \label{fig:training}
\end{figure}
%

%% file: sec/4_evaluation.tex
\section{Evaluation}

\subsection{Metrics}
\label{sec:evaluation:metrics}
The RMSE quantifies the overall deviation by computing the square root of the mean squared Euclidean distances between the predicted keypoints and the ground truth keypoints~\cite{Mathis:2018}.
The Percentage of Correct Keypoints (PCK) metric quantifies the proportion of predicted keypoints that fall within a specified threshold~\cite{Badger:2020}.
PCK05 and PCK10 are computed, where the threshold is a fraction ($0.05$ and $0.1$, respectively) of the largest dimension of the ground truth bounding box in 2D and the maximum distance between any two keypoints in 3D.

The RMSE offers a global measure of accuracy by quantifying the average error across all keypoints and is sensitive to outliers.
PCK on the other hand, focuses on the proportion of keypoints that meet a predefined accuracy threshold, making it less sensitive to outliers and more dependent on the size and scale of the tracked subject.
Thus, PCK can be seen as the more meaningful performance measure for pose estimation tasks~\cite{Waldmann:2024}.

\subsection{Results}
\label{sec:evaluation:results}
\input{tables/Results_3D-POP_2D}
\input{tables/Results_3D-POP_3D}
\input{tables/Results_Wild-MuPPET}

%
%
2D results on 3D-POP are in~\cref{tab:2D}.
DLCISO
performs better than DLCSAM in all metrics.
However, 3D-MuPPET~\cite{Waldmann:2024} achieves more accurate results with a median of $4.7$ px (vs. $7.6$ for DLCISO).
%
3D pose estimation results can be found in~\cref{tab:3D}.
While 3D-DLCISO performs better than 3D-DLCSAM across most metrics, the latter achieves the lowest RMSE of $20.5$ mm.
%
This hints at a reduced number of large outliers compared to the other two pose estimators.
%
%
%
In~\cref{tab:Wild} we report
3D results on Wild-MuPPET.
%
%
DLCISO-YOLO outperforms DLCISO across all metrics.
%
While~\cite{Waldmann:2024} is the most accurate model across most metrics, DLCISO-YOLO has the lowest RMSE with $32.1$ mm (vs. $53.5$ mm for~\cite{Waldmann:2024}).
This RMSE together with a slightly higher median ($15.0$ mm for~\cite{Waldmann:2024} vs. $21.5$ mm for DLCISO-YOLO) hints at a decrease in the occurrence of significant outliers compared to 3D-MuPPET~\cite{Waldmann:2024}.
%
%
%
Overall, the results highlight that our texture-independent approach is comparable to the texture-dependent framework~\cite{Waldmann:2024} with a RMSE of $20.5$ mm (vs. $25.0$ mm) on 3D-POP and $32.1$ mm (vs. $53.4$ mm) on Wild-MuPPET.
%

\subsection{Ablation Studies}
\subheading{Outlier Analysis}
\input{tables/Results_Ablation}
For further insights into the limitations of our model, we conduct a keypoint-wise analysis using DLCISO on the 3D-POP dataset~\cite{Naik:2023}. The results are summarized in~\cref{tab:Ablation}.
Keypoints located on or near the head (e.g., beak, nose, eye(l), and eye(r)) consistently show the lowest median errors and the highest PCK scores. This suggests that head keypoints are more reliably localized, likely due to their distinctive shape and higher visibility in silhouette data. 
In contrast, keypoints positioned further along the body show a lower accuracy. While most of the body keypoints perform moderately well at the PCK10 metric (over $70\%$), they drop significantly under the stricter PCK05 metric, suggesting they are detected with reasonable proximity but not high precision. These areas are more prone to pose variability, making accurate prediction from silhouettes particularly challenging. The tail exhibits the poorest performance with a median error of $21.0$ mm and a very low PCK05 of only $1.9\%$, highlighting it as a major source of error.

\subheading{DLC}
To assess the impact of data augmentation on model performance, we retrained DLCISO  using a modified augmentation strategy. Specifically, we disabled augmentations with no clear benefit for silhouette-based pose estimation (contrast variation, Gaussian noise, and motion blur) and enabled left-right flipping. All other training parameters and data were kept identical.
Compared to the original configuration, the augmented model shows slightly degraded performance across all metrics (RMSE: $21.1$ mm vs. $22.2$ mm, Median: $13.5$ mm vs. $14.1$ mm, PCK05: $35.4\%$ vs. $33.3\%$, PCK10: $74.5\%$ vs. $71.4\%$).

\subheading{SAM}
\input{tables/Results_Sam}
To investigate the influence of segmentation quality on pose estimation performance, we evaluated DLCISO using silhouettes generated from different SAM configurations: the huge and base variants, as well as a fine-tuned version of the base model. We freeze the encoders for fine-tuning
and adapt only the mask decoder weights.
Results are summarized in~\cref{tab:Sam}. The SAM-huge model yields the best overall performance (Median: $13.5$ mm, PCK10: $74.5\%$). Fine-tuning the SAM-base model reduces RMSE ($23.9$ vs. $20.6$ mm) and improves PCK10 ($68.0$ vs. $71.3\%$),
indicating only a limited benefit from domain-specific fine-tuning in this setting.

\subsection{Species Transfer}
\input{tables/Results_species_transfer}
To demonstrate the applicability of the model to other bird species without requiring additional training or fine-tuning,
we estimate 2D keypoints on four species of the Animal Kingdom dataset~\cite{Ng:2022}:
pigeons, stock doves, turtle doves and ravens.
For the first three, we evaluate on all images.
%
Results on six keypoints (beak, nose, eyes(l/r), shoulders(l/r), tail) are in~\cref{tab:species-transfer}.
%
From this evaluation we notice that most images display different angles and behaviour (e.g.flying behaviour filmed from far away) than 3D-POP~\cite{Naik:2023}.
That is why we also evaluate on $73$ manually selected images of single ravens which display a similar angle and behaviour to 3D-POP.
We choose ravens
because ravens are similar in size as a pigeon, and their images in Animal Kingdom~\cite{Ng:2022} have a similar posture and viewing angle to those from 3D-POP~\cite{Naik:2023}.
%
%
For ravens we achieve better results with a PCK10 of $51.9\%$ compared to the other three bird species (e.g. PCK10 of $25.5\%$ for stock dove). 
This indicates that the framework struggles to predict new unseen poses and angles but is capable of generalizing to new species with preliminary promising results.

%% file: tables/Results_3D-POP_2D.tex
\begin{table}
    \centering
    \begin{tabular}{lcaa}
    \toprule
    Metric/Method & DLC*~\cite{Waldmann:2024} & DLCSAM & DLCISO\\
    \midrule
    RMSE (px)$\downarrow$   & $\mathbf{39.0}$ & $49.1$ & $41.7$\\
    Median (px)$\downarrow$ & $\mathbf{4.7}$  &  $8.3$ & $7.6$ \\
    PCK05 (\%)$\uparrow$    & $\mathbf{89.1}$ & $67.3$ & $69.0$\\
    PCK10 (\%)$\uparrow$    & $\mathbf{96.8}$ & $88.7$ & $89.5$\\
    \bottomrule
    \end{tabular}
    \caption{
    \textit{Quantitative 2D results on 3D-POP.}
    We report the RMSE and its median (px), PCK05 (\%) and PCK10 (\%)
    on the four test sequences, cf.~\cref{sec:framework:datasets}.
    %
    Comparison 
    with~\cite{Waldmann:2024} (ours highlighted in gray).
    %
    %
    Upward and downward arrows indicate whether a higher or lower value is strived for.
    Best results per row in bold.}
    \label{tab:2D}
\end{table}

%% file: tables/Results_3D-POP_3D.tex
\begin{table}
    \centering
    {\setlength{\tabcolsep}{1.5pt}
    \begin{tabular}{lcaa}
    \toprule
    Metric/Method & 3D-DLC*~\cite{Waldmann:2024} & 3D-DLCSAM & 3D-DLCISO\\
    \midrule
    RMSE (mm)$\downarrow$   & $25.0$          & $\mathbf{20.5}$ & $21.1$\\
    Median (mm)$\downarrow$ & $\mathbf{7.5}$  & $14.2$          & $13.5$  \\
    PCK05 (\%)$\uparrow$    & $\mathbf{66.1}$ & $33.9$          & $35.4$\\
    PCK10 (\%)$\uparrow$    & $\mathbf{90.9}$ & $73.6$          & $74.5$\\
    \bottomrule
    \end{tabular}}
    \caption{\textit{Quantitative 3D results on 3D-POP.}
    We report the RMSE and its median (mm), PCK05 (\%) and PCK10 (\%)
    on the four test sequences (see \cref{sec:framework:datasets}).
    %
    Comparison
    with~\cite{Waldmann:2024} (ours highlighted in gray).
    %
    %
    Upward and downward arrows indicate whether a higher or lower value is strived for.
    Best results per row in bold.}
    \label{tab:3D}
\end{table}

%% file: tables/Results_Wild-MuPPET.tex
\begin{table}
    \centering
    {\setlength{\tabcolsep}{1.5pt}
    \begin{tabular}{lcaa}
    \toprule
    Metric/Method & Wild-DLC~\cite{Waldmann:2024} & DLCISO & DLCISO-YOLO\\ \midrule
    RMSE (mm)$\downarrow$   & $53.4$          & $44.3$ & $\mathbf{32.1}$\\
    Median (mm)$\downarrow$ & $\mathbf{15.0}$ & $25.3$ & $21.5$  \\
    PCK05 (\%)$\uparrow$    & $\mathbf{25.1}$ & $7.3$  & $8.8$\\
    PCK10 (\%)$\uparrow$    & $\mathbf{74.4}$ & $39.2$ & $48.3$\\
    \bottomrule
    \end{tabular}}
    \caption{\textit{3D results on Wild-MuPPET.}
    We report the RMSE and its median (mm), PCK05 (\%) and PCK10 (\%)
    on the test sequences.
    %
    Comparison
    with~\cite{Waldmann:2024} (ours highlighted in gray).
    %
    %
    Upward and downward arrows indicate whether a higher or lower value is better.
    Best results per row in bold.}
    \label{tab:Wild}
\end{table}
%

%% file: tables/Results_Ablation.tex
\begin{table*}
    \centering
    \begin{tabular}{lccccccccc}
    \toprule
    Metric/Keypoint         & Beak   & Nose   & Eye(l) & Eye(r) & Shoulder(l) & Shoulder(r) & Keel(t) & Keel(b) & Tail \\
    \midrule
    RMSE (mm)$\downarrow$   & $25.1$ & $23.3$ & $25.0$ & $23.9$ & $16.6$ & $17.4$ & $18.1$ & $19.1$ & $25.7$\\
    Median (mm)$\downarrow$ &  $7.0$ &  $6.7$ &  $6.9$ &  $7.1$ & $12.3$ & $14.1$ & $16.1$ & $15.9$ & $21.0$ \\
    PCK05 (\%)$\uparrow$    & $63.1$ & $72.0$ & $69.7$ & $70.0$ & $35.9$ & $31.6$ & $20.3$ & $13.6$ &  $1.9$\\
    PCK10 (\%)$\uparrow$    & $80.3$ & $88.3$ & $89.9$ & $86.9$ & $82.2$ & $72.6$ & $74.7$ & $71.0$ & $44.0$\\
    \bottomrule
    \end{tabular}
    \caption{
    \textit{Comparison between different keypoints.}
    %
    We show results for the DLCISO model
    for each keypoint.
    We report the RMSE and its median (mm), PCK05 (\%) and PCK10 (\%)
    on the four test sequences, cf.~\cref{sec:framework:datasets}.
    Upward and downward arrows indicate whether a higher or lower value is strived for.}
    \label{tab:Ablation}
\end{table*}

%% file: tables/Results_Sam.tex
\begin{table}
    \centering
    \begin{tabular}{lccc}
    \toprule
    Method        & DLCISO & DLCISO & DLCISO\\
    SAM-Type      & Huge   & Base   & Base (Finetuned)\\
    \midrule
    RMSE (mm)$\downarrow$   & $21.1$ & $23.9$ & $20.6$\\
    Median (mm)$\downarrow$ & $13.5$ & $13.7$ & $14.5$\\
    PCK05 (\%)$\uparrow$    & $35.4$ & $37.0$ & $31.8$\\
    PCK10 (\%)$\uparrow$    & $74.5$ & $68.0$ & $71.3$\\
    \bottomrule
    \end{tabular}
    \caption{
    \textit{Comparison between different SAM configurations.}
    %
    We report the RMSE and its median (mm), PCK05 (\%) and PCK10 (\%)
    on the four test sequences, cf.~\cref{sec:framework:datasets}.
    Upward and downward arrows indicate whether a higher or lower value is strived for.}
    \label{tab:Sam}
\end{table}

%% file: tables/Results_species_transfer.tex
\begin{table}
    \setlength{\tabcolsep}{3pt}
    \centering
    \begin{tabular}{@{}lcccc@{}}
    \toprule
    Metric/Species & Pigeon & Stock Dove & Turtle Dove & Raven\\ \midrule
    RMSE (px)$\downarrow$   & $107.6$ & $125.6$ & $90.3$ & $50.2$\\
    Median (px)$\downarrow$ & $45.0$  & $34.8$  & $23.0$ & $27.8$\\
    PCK05 (\%)$\uparrow$    & $10.2$  & $8.5$   & $9.4$  & $14.0$\\
    PCK10 (\%)$\uparrow$    & $19.9$  & $25.5$  & $24.0$ & $51.9$\\
    \bottomrule
    \end{tabular}
    \caption{\textit{
    Species transfer.}
    We report the RMSE and its median (px), PCK05 (\%) and PCK10 (\%)
    on four different species from Animal Kingdom~\cite{Ng:2022}.
    We use DLCISO trained on texture-independent pigeon data from 3D-POP only.
    %
    %
    }
    \label{tab:species-transfer}
\end{table}

%% file: sec/5_conclusion.tex
\section{Limitations and Future Work}
While our framework shows promising performance given its reliance on silhouette data, the overall accuracy, particularly in terms of RMSE and PCK, still needs improvement for downstream applications.
%
%
%
%
%
One limitation is the quality of the silhouettes, which can be improved.
%
One approach here is to fine-tune the SAM-huge model~\cite{Kirillov:2023} specifically for this domain.
%
%
Another limitation of the 2D pose estimation on silhouettes is the potential confusion of left and right keypoints, such as shoulders and eyes.
In silhouette data, it is often ambiguous which direction an individual faces, making it difficult to differentiate between the left and right sides.
A possible solution could be to triangulate different combinations of left and right keypoints and subsequently select the combination with the smallest triangulation error.
Another solution is to leverage texture and geometry features instead of relying on silhouettes only. Specifically, enhancing silhouettes with internal features (e.g., contours generated by a Canny edge detector) to provide additional cues for pose estimation.
To reduce the number of keypoint outliers, we will leverage temporal information within our framework.
In addition, we aim at a comparison with UniAP~\cite{Sun:2024} on the bird instances included in Animal Kingdom~\cite{Ng:2022}.
We will release code once the project is complete.

\section{Conclusion}
%
We present an extension to the 3D-MuPPET framework~\cite{Waldmann:2024}, transforming it into a texture-independent framework.
%
We demonstrate that
the extension maintains its capability to reliably track and predict the 2D and 3D poses of multiple pigeons, with a performance
comparable to that of the original 3D-MuPPET~\cite{Waldmann:2024}.
%
Additionally, the proposed approach shows potential for the application to other phenotypes and species.
%
Our preliminary results on four bird species are promising.
To achieve reliable results, further work and optimization is necessary.
However, our approach establishes a promising foundation for the development of more robust and accurate texture-independent pose estimation methods that can be applied to new species.

%% file: sec/6_acknowledgements.tex
\section*{Acknowledgments}
We thank Leon Wenzler for his support in deploying our training scripts on the Collective Computational Unit.

Funded by the Deutsche Forschungsgemeinschaft (DFG, German Research Foundation) under Germany's Excellence Strategy – EXC 2117 – 422037984, and the Federal Ministry of Education and Research (BMBF) within the research program KI4KMU under grant number 01IS23046B (ARGUS).
U.W. acknowledges funding from the Connected Minds Program, supported by Canada First Research Excellence Fund, Grant \#CFREF-2022-00010.